\documentclass[sigconf]{acmart}
\usepackage{listings}
\lstdefinelanguage{json}{
    basicstyle=\normalfont\ttfamily,
    numbers=left,
    numberstyle=\tiny,
    stepnumber=1,
    numbersep=5pt,
    showstringspaces=false,
    breaklines=true,
    frame=lines,
    backgroundcolor=\color{gray!10},
    literate=
     *{0}{{{\color{black}0}}}{1}
      {1}{{{\color{black}1}}}{1}
      {2}{{{\color{black}2}}}{1}
      {3}{{{\color{black}3}}}{1}
      {4}{{{\color{black}4}}}{1}
      {5}{{{\color{black}5}}}{1}
      {6}{{{\color{black}6}}}{1}
      {7}{{{\color{black}7}}}{1}
      {8}{{{\color{black}8}}}{1}
      {9}{{{\color{black}9}}}{1}
      {:}{{{\color{black}:}}}{1}
      {,}{{{\color{black},}}}{1}
      {"}{{{\color{blue}"}}}{1},
}
\usepackage{subcaption}
\usepackage{fontawesome5}
\usepackage{url}
\AtBeginDocument{%
  }

\setcopyright{cc}
\copyrightyear{2025}
\acmYear{2025}
\acmDOI{}

\acmConference[Agentic \& GenAI Evaluation Workshop KDD'25]{KDD workshop on Evaluation and Trustworthiness of Agentic and Generative AI Models}{August 4, 2025}{Toronto, ON, Canada}

\acmISBN{}

\begin{document}

\title[]{DialogueForge: LLM Simulation of Human-Chatbot Dialogue}

\author{Ruizhe Zhu}
\authornote{Equal contribution.}
\affiliation{
  \institution{ETH Zurich}
  \city{Zurich}
  \state{}
  \country{Switzerland}
}

\author{Hao Zhu}
\authornotemark[1]
\affiliation{%
  \institution{ETH Zurich}
  \city{Zurich}
  \state{}
  \country{Switzerland}
}

\author{Yaxuan Li}
\authornotemark[1]
\affiliation{%
  \institution{ETH Zurich}
  \city{Zurich}
  \state{}
  \country{Switzerland}
}

\author{Syang Zhou}
\affiliation{%
  \institution{Calvin Risk AG}
  \city{Zurich}
  \state{}
  \country{Switzerland}
}

\author{Shijing Cai}
\affiliation{%
  \institution{Calvin Risk AG}
  \city{Zurich}
  \state{}
  \country{Switzerland}
}
\author{Ma\l{}gorzata \L{}azuka}
\orcid{0009-0006-6261-6136}
\affiliation{%
  \institution{Calvin Risk AG}
  \city{Zurich}
  \state{}
  \country{Switzerland}
}

\author{Elliott Ash}
\orcid{0000-0002-6817-7529}
\affiliation{%
  \institution{ETH AI Center, ETH Zurich}
  \city{Zurich}
  \state{}
  \country{Switzerland}
}

\renewcommand{\shortauthors}{R. Zhu, H. Zhu, Y. Li, S. Zhou, S. Cai, M. \L{}azuka, and E. Ash}

\begin{abstract}
  Collecting human-chatbot dialogues typically demands substantial manual effort and is time-consuming, which limits and poses challenges for research on conversational AI. In this work, we propose DialogueForge -- a framework for generating AI-simulated conversations in human-chatbot style. To initialize each generated conversation, DialogueForge uses seed prompts extracted from real human-chatbot interactions. We test a variety of LLMs to simulate the human chatbot user, ranging from state-of-the-art proprietary models to small-scale open-source LLMs, and generate multi-turn dialogues tailored to specific tasks. In addition, we explore fine-tuning techniques to enhance the ability of smaller models to produce indistinguishable human-like dialogues. We evaluate the quality of the simulated conversations and compare different models using the UniEval and GTEval evaluation protocols. Our experiments show that large proprietary models (e.g., GPT-4o) generally outperform others in generating more realistic dialogues, while smaller open-source models (e.g., Llama, Mistral) offer promising performance with greater customization. We demonstrate that the performance of smaller models can be significantly improved by employing supervised fine-tuning techniques. Nevertheless, maintaining coherent and natural long-form human-like dialogues remains a common challenge across all models.
\end{abstract}


\newcommand\blfootnote[1]{%
  \begingroup
  \renewcommand\thefootnote{}\footnote{#1}%
  \addtocounter{footnote}{-1}%
  \endgroup
}

\maketitle

\blfootnote{\\Author contributions using the CRediT framework \cite{brand2015beyond, devos2024few}:\\
Ruizhe Zhu: Dialogue Generation, Experiments, Codebase maintenance, Writing\\
Hao Zhu: Dialogue Evaluation, Visualization, Writing\\
Yaxuan Li: Datasets, Experiments, Writing\\
Syang Zhou, Shijing Cai, Ma\l{}gorzata \L{}azuka: Conceptualization, Data Preprocessing, Supervision, Financial Support, Review\\
Elliott Ash: Supervision\\
\faEnvelopeOpen[regular]\ \ Syang Zhou:\ \href{mailto:sz@calvin-risk.com}{sz@calvin-risk.com}
} 

\section{Introduction}
Large language models (LLMs) have rapidly emerged as a transformative force in artificial intelligence, which marks a significant milestone in the development of natural language processing (NLP), driving substantial progress across a wide range of NLP tasks and paving new paths for research. Their ability to generate coherent and human-like text has made them central to the development of advanced dialogue systems~\cite{deng-etal-2023-prompting}. As these models become more capable, the demand for high-quality human-chatbot dialogue data, both for training and evaluation, continues to grow~\cite{soudani2024surveyrecentadvancesconversational}.

In order to advance the study and development of LLMs, large-scale human-chatbot dialogue data collection is indispensable. However, gathering and annotating these human-involved dialogues remains a major bottleneck. Traditional crowd-sourcing data collection methods, which rely on human participants and manual annotation, are time-consuming, costly, and often difficult to scale, which can pose significant challenges to the research process~\cite{10.1145/3589335.3641238}. 
LLMs have recently emerged as effective solutions to this challenge. Their powerful architectures and capacity to learn from massive datasets allow them to produce contextually appropriate, informative, and human-like responses~\cite{10.1145/3589335.3641238}. 
Several recent works leverage large language models directly for synthetic dialogue generation:

\textit{One-Go In-Context Prompting.}
A pre-trained LLM is prompted with a seed (topic description, knowledge-graph-derived summary, or few-shot examples) to produce an entire multi-turn conversation in a single pass. For example, PLACES~\cite{chen-etal-2023-places} uses few-shot prompts combining human-written background info and sample turns to GPT-3 to generate full dialogues. SODA~\cite{kim-etal-2023-soda} employs GPT-3.5 to expand knowledge graph triplets into short narrative seeds and then into multi-turn chats. BotChat~\cite{duan2023botchatevaluatingllmscapabilities} leverages multiple different LLMs to generate multi-turn dialogues utterance-by-utterance from authentic human-bot chat seeds.

\textit{Fine-Tuning.} The LLM is first fine-tuned on a small dialogue completion corpus, then used for generation, for instance, AUGESC~\cite{zheng2023augescdialogueaugmentationlarge} fine-tunes GPT-J on ESConv~\cite{liu-etal-2021-towards} dataset, then generates conversations from a description plus the first turn.

\textit{Turn-by-Turn Multi-Agent Simulation.} Two or more LLM agents converse sequentially, often to model different personas or to mix skills: PERSONACHATGEN~\cite{lee-etal-2022-personachatgen} runs two GPT-3 instances, each conditioned on distinct persona profiles, generating one turn at a time. BOTSTALK~\cite{kim-etal-2022-botstalk} engages multiple GPT-3 models that alternate turns, selecting from different skill-specific datasets turn by turn.

\emph{Task-Oriented In-Context LLM Simulators.} Prompt-based LLM methods generate dialogues for task-completion settings without any fine-tuning. ICL-US~\cite{tseng-etal-2021-transferable} uses few-shot examples plus a user goal and history to generate user-agent turns via in-context learning. Dialogic~\cite{li-etal-2022-controllable} similarly prompts GPT-3 with ontology-extracted goals and in-context examples, then applies a critic step to enforce belief-state alignment.

While prior LLM-based conversation generation techniques tend to specialize in a single paradigm, whether one-go in-context prompting, fine-tuning, or turn-by-turn multi-agent simulation, we introduce {\textbf{DialogueForge}}, a novel framework that unifies these strategies. DialogueForge starts from authentic human-chatbot exchanges, using only the first user utterance as a prompt. It infers each dialogue’s underlying task objective from the original conversation and then iteratively generates alternating “Inquirer” (human) and “Responder” (bot) turns with a chosen LLM (see Figure \ref{fig:chatbot}). This approach enables scalable synthesis of diverse, task-tailored multi-turn dialogues, drastically reducing cost and time required to collect human-chatbot style conversation data manually.

To assess LLMs’ ability to produce human-like conversational flow, adhere to task goals, and sustain coherent dialogue across multiple turns, DialogueForge adopts two evaluation metrics proposed in~\cite{duan2023botchatevaluatingllmscapabilities}. Both metrics employ an LLM-as-a-judge approach: \textbf{UniEval} evaluates individual dialogue quality, while \textbf{GTEval} compares generated conversations against ground-truth conversations.

Finally, to further boost generation quality, DialogueForge applies supervised fine-tuning: base LLMs are trained on a curated corpus of high-quality human-chatbot exchanges that span varied tasks, linguistic styles, and interaction patterns. Through this fine-tuning process, the model learns to predict each next turn given full dialogue context, which reinforces realistic conversational behaviors and enhances the overall naturalness of synthetic dialogues.

In this work, we make three key contributions:

\begin{enumerate}
    \item \textbf{DialogueForge}, a unified synthetic conversation generation framework that seeds real human-chatbot exchanges with only the first user utterance, automatically infers each dialogue’s task goal, and combines in-context LLM prompting, iterative multi-agent simulation, and supervised fine-tuning to produce richly varied, task-tailored multi-turn dialogues at scale.
    \item A comprehensive evaluation of both proprietary (e.g. GPT-4o) and open-source (e.g., Llama, Mistral) LLMs using two LLM-as-judge metrics: UniEval and GTEval, showing that (a) larger models generally outperform smaller ones, (b) dialogue quality degrades as conversation length increases, and (c) fine-tuning substantially boosts the performance of smaller models.
    \item An analysis of evaluation bias, where we adapt BotChat’s judging protocols~\cite{kim-etal-2022-botstalk} to compare different judge LLMs (GPT-4o, Claude 3.7, Gemini 2.0 Flash), demonstrating the robustness of our metrics across judge choices.
\end{enumerate}
All code, models, and datasets are publicly available on GitHub\footnote{Available at \url{https://github.com/nerchio/Human_Chatbot-Generation}\label{repo}}. 
We believe that developing DialogueForge and releasing open-source code of the framework as well as conversation data and fine-tuned LLMs, we will enable and accelerate further research in the area of conversational AI, allowing researchers with limited resources to contribute to this field. 
Our approach addresses a key challenge faced in dialogue research: the expensive and time-consuming process of collecting human-chatbot interactions, by providing a unified framework that can generate diverse, task-oriented conversations at scale. 
The demonstrated performance of fine-tuned smaller models suggests that high-quality dialogue generation does not necessarily require large proprietary systems, which has important implications for research accessibility and the development of conversational AI in resource-limited settings.

The rest of the paper is organized as follows: Section 2 reviews related work on LLM-based dialogue generation and evaluation; Section 3 details the DialogueForge methodology, including seed prompt extraction, dialogue generation, and fine-tuning strategies; Section 4 introduces our evaluation setup (Section 4.1) and presents experimental results (Section 4.2); Section 5 discusses the implications, limitations, and potential enhancements of DialogueForge; and Section 6 concludes and outlines directions for future work.

\section{Related Works}
\subsection{LLM Data Generation}
Recent advances in synthetic dialogue generation have been driven by large-scale, high-quality datasets. SODA~\cite{kim-etal-2023-soda} distills socially contextualized dialogues from LLMs, producing more natural and consistent conversations. UltraChat~\cite{ding-etal-2023-enhancing} scales instruction-tuned dialogue generation with 1.5M synthetic conversations, enabling the strong UltraLM model. Baize~\cite{xu-etal-2023-baize} generates self-chat data using ChatGPT and introduces feedback-based self-distillation to fine-tune open-source models. Complementing synthetic data, WildChat~\cite{zhao2024wildchat1mchatgptinteraction} provides a real-world dataset of 1M ChatGPT-user interactions, enriched with demographic metadata. Lately, LLM Roleplay~\cite{tamoyan2024llmroleplaysimulatinghumanchatbot} demonstrated that persona-based conditioning improves engagement and specificity. These works collectively highlight trends in scaling, personalization, and leveraging both synthetic and real-world data to improve dialogue quality. 

Recently, numerous studies have employed fine-tuning techniques spanning full fine-tuning, adapter-based PEFT, and progressive learning to improve the models' ability to generate high-quality data following specific instructions. For example, the Falcon series~\cite{almazrouei2023falconseriesopenlanguage} fine‐tunes open LLMs on massive, diverse text corpora to boost generation fluency and factual consistency. More recently, Orca~\cite{mukherjee2023orcaprogressivelearningcomplex} employs progressive fine-tuning on GPT-4 explanation traces to teach complex reasoning patterns, and WizardLM~\cite{xu2025wizardlmempoweringlargepretrained} fine-tunes large pre-trained models to follow intricate multi-step instructions with high fidelity.

\subsection{Human-Chatbot Datasets}
Several large-scale conversational datasets have been released to support both open-domain chit-chat and human-chatbot interaction research. PersonaChat~\cite{zhang-etal-2018-personalizing} comprises crowd-authored daily chit-chat dialogues between paired personas. The OpenAssistant~\cite{10.5555/3666122.3668186} dataset provides human-crafted assistant responses to a wide variety of prompts in English, as do Anthropic HH~\cite{ganguli2022redteaminglanguagemodels}, Chatbot Arena~\cite{10.5555/3692070.3692401}, and LMSYS-Chat-1M~\cite{zheng2024lmsyschat1mlargescalerealworldllm}, each collecting high-quality human-bot exchanges across diverse topics. Finally, MT Bench~\cite{bai-etal-2024-mt} contributes fine-grained human judgments on model outputs to benchmark machine translation in conversational settings. Together, these resources underpin much of the recent progress in training and evaluating advanced dialogue systems.

\subsection{Evaluation of LLMs}
To better assess LLM competence, \citet{hendrycks2021measuringmassivemultitasklanguage} introduced the MMLU benchmark, which tests multitask language understanding across 57 diverse subjects. Addressing the challenge of multi-step reasoning, particularly in math, \citet{cobbe2021trainingverifierssolvemath} proposed a verifier-based method to improve model performance on the GSM8K dataset.  Meanwhile, GPTScore~\cite{fu2023gptscoreevaluatedesire} is a customizable, model-based evaluation framework that leverages large pre-trained language models for multi-dimensional text assessment without requiring annotated data. \citet{duan2023botchatevaluatingllmscapabilities} used three evaluation protocols to assess LLM's capability of having multi-turn dialogues. These efforts collectively contribute to both developing more capable models and building the tools necessary to assess their real-world performance more effectively.  

\begin{figure}[t]
\includegraphics[width=\columnwidth]{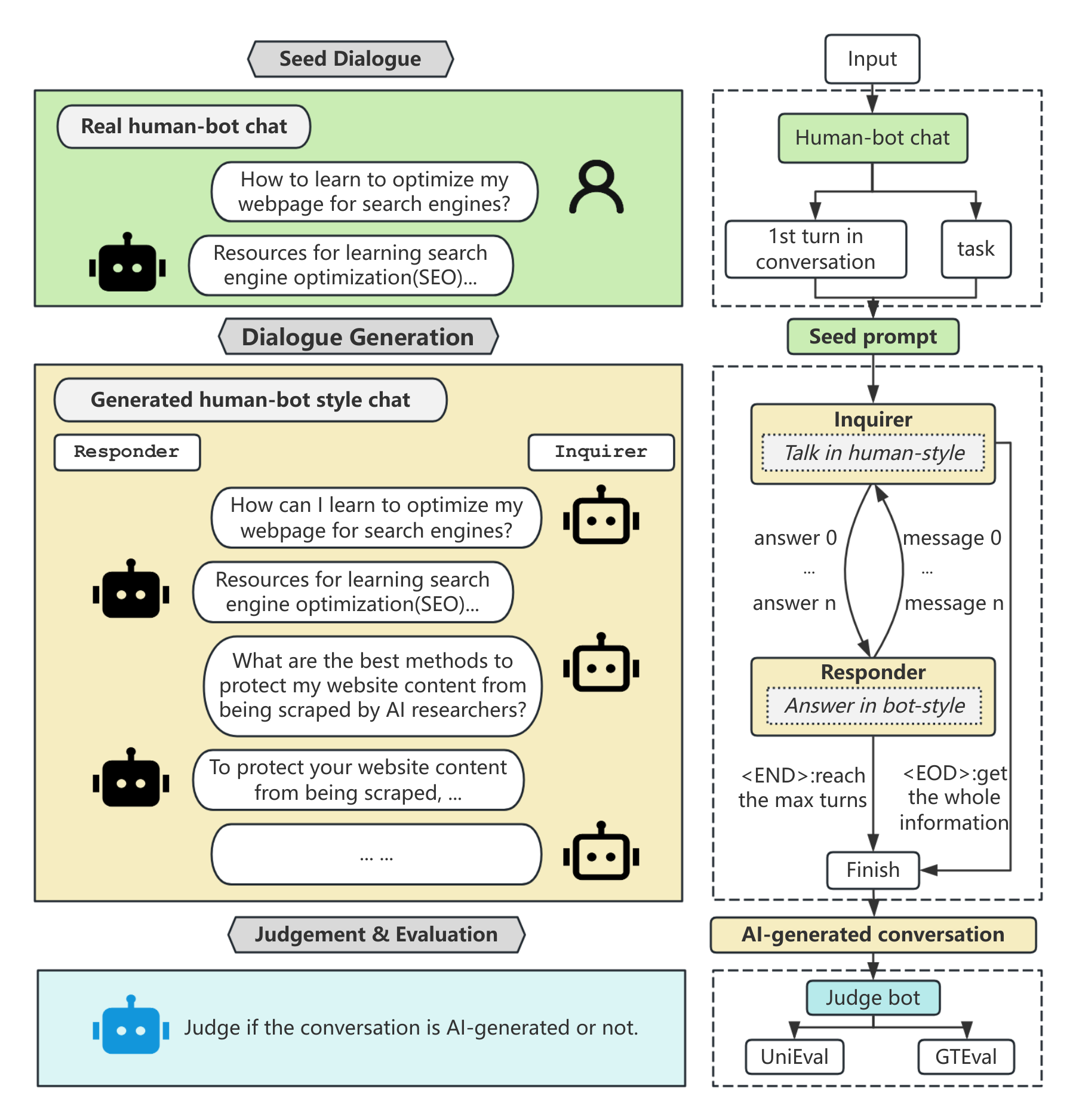}
  \caption{Flow of the dialogue generation and evaluation process of DialogueForge.}
  \label{fig:chatbot}
  \Description{1. Seed dialogue is sourced from a real human-bot chat and used as the first turn in conversation, determining the task. 2. The seed prompt is used for dialogue generation, where an inquirer (LLM instructed to talk in human-style) and responder (LLM instructed to answer in bot-style) take turns exchanging messages. 3. The AI-generated conversation is passed to the judge bot, which assesses the generated dialogue using the UniEval and GTEval metrics.}
\end{figure}

\section{Methodology}
\label{sec:methodology}

In this section, we present the general structure of DialogueForge, our dialogue generation methodology, and further implementation details. First, Section~\ref{subsec:architecture} presents a high-level overview of the dialogue generation and evaluation workflow used in DialogueForge. Then, Section~\ref{subsec:implementation} provides a detailed description of DialogueForge implementation, including the LLMs used internally to generate chat messages. Finally, Section~\ref{subsec:fine-tuning} discusses our experimental methodology of fine-tuning smaller LLMs to enhance their performance in generating human-like conversations.

\subsection{DialogueForge Architecture}
\label{subsec:architecture}

The overall process of dialogue generation and evaluation implemented in DialogueForge has been adapted from BotChat~\cite{duan2023botchatevaluatingllmscapabilities} and is illustrated in Figure~\ref{fig:chatbot}.

\subsubsection{Seed prompt extraction}
We begin by seeding the system with a real dialogue segment, extracted from an actual interaction between a human and a chatbot. 
This seed consists of an initial human query and the corresponding response from the chatbot. 
Additionally, the seed prompt extraction process involves parsing the original human-chatbot exchange to identify the central topic of the conversation, which is then embedded into generation prompts.
This topic serves as a guiding constraint, helping to ensure that the subsequent generated dialogue remains coherent and does not deviate significantly from the theme of the original exchange.

However, it is important to note that DialogueForge does not attempt to faithfully recreate the original seed dialogue beyond utilizing the initial utterance as a generation prompt, nor does it explicitly evaluate task completion, toxicity or hallucination rates. 
The framework generates entirely novel conversational trajectories guided by the inferred topic. 
Our evaluation methodology focuses exclusively on conversational naturalness and the ability to produce human-indistinguishable dialogue, as quantified through UniEval and GTEval metrics. 
This design choice reflects our primary research objective of advancing synthetic dialogue generation quality rather than task-oriented performance assessment.

\subsubsection{Dialogue generation}
Following this initialization, we simulate an extended multi-turn dialogue by alternating between two large language models, each assuming a distinct conversational role. The first model, referred to as the \textbf{inquirer}, acts as the simulated human participant, while the other model, called the \textbf{responder}, plays the role of the chatbot. 
Importantly, we treat the inquirer and the responder models as two independent agents. They are instantiated with different prompts and can even be based on entirely different LLM architectures. 
This allows us to study agent interactions with a high degree of flexibility and realism. 

Throughout the generation process, the full history of the conversation is passed between the two models at each turn, enabling context-aware responses. The generation continues iteratively, with each model producing one utterance per turn, until either (a) a predefined maximum number of turns is reached, or (b) the inquirer determines that the dialogue has naturally concluded. In our experimental setup, we set strict turn limits of 6 and 12 turns to enable controlled comparison across models and datasets. The inquirer model is therefore not explicitly programmed to recognize task completion, and the termination is mainly based on these predetermined constraints.

\subsubsection{Judgement}
The AI-generated conversation between the inquirer and the responder models generated in the dialogue generation step is then passed to the judge bot. The judge assesses the dialogue using UniEval and GTEval metrics (discussed in more detail in Section~\ref{subsec:metrics}).

\subsection{Implementation}
\label{subsec:implementation}

The implementation of DialogueForge leverages LangGraph~\cite{langgraph}, a framework designed to streamline the construction of multi-agent communication pipelines. LangGraph facilitates the orchestration of message passing between agents and provides mechanisms for specifying various termination conditions, enabling more structured and manageable dialogue simulations. Below, we provide further details regarding the real-world dialogue datasets and large language models used in DialogueForge.

\subsubsection{Seed dialogue datasets}
\label{subsubsec:seed_datasets}
Obtaining high-quality, large-scale public datasets containing real human-chatbot dialogues remains a non-trivial challenge due to the difficulty of large-scale data collection and privacy concerns. In this study, we leverage two publicly available datasets hosted on Hugging Face: OpenAssistant Conversations (OASST1)~\cite{10.5555/3666122.3668186} with 413 conversations and Chatbot Arena Conversations~\cite{arena} with 1,002 conversations.
To ensure consistency and relevance for our experiments, we performed data cleaning and reformatting to extract coherent and contextually rich human-chatbot dialogue samples. These curated datasets serve as the basis for all subsequent evaluations. More details on data preprocessing can be found in our publicly available code repository\footref{repo}.

To investigate how dialogue length affects model behavior and performance, we define two dialogue length constraints in our experimental design: a maximum of 6 turns and a maximum of 12 turns per conversation. This stratification enables us to analyze model robustness and consistency across varying interaction depths.

\subsubsection{Inquirer LLMs}
The primary focus of our research is to improve the performance of the inquirer agent (that is, the model that drives the conversation and simulates the human chatbot user). To that end, we consider and evaluate a variety of candidate models for this role, grouped into two broad categories:
\begin{itemize}
    \item Large, high-capacity LLMs that demonstrate strong conversational fidelity and produce human-like dialogue, but are often resource-intensive and impractical for local or edge deployments:
    \begin{itemize}
        \item GPT-4o \cite{openai2024gpt4ocard}
        \item GPT-4o mini \cite{openai2024gpt4ocard}
        \item DeepSeek-V3 \cite{deepseek}
        \item Llama-3.3-70B \cite{grattafiori2024llama3herdmodels}
        \item Gemma-2-27B \cite{gemma}
    \end{itemize}
    \item Smaller, lightweight models that are more computationally efficient and better suited for deployment in resource-constrained environments, yet may underperform in generating natural and engaging conversations:
    \begin{itemize}
        \item Llama-3.2-3B \cite{grattafiori2024llama3herdmodels}
        \item Llama-3.1-8B \cite{grattafiori2024llama3herdmodels}
        \item Mistral-7B \cite{jiang2023mistral7b}
    \end{itemize}
\end{itemize}
To bridge the performance gap between large and small models, we apply supervised fine-tuning to the smaller models, described in detail in Section~\ref{subsec:fine-tuning}.

\subsubsection{Responder LLM}
For all experiments, we designate GPT-4o mini as the responder model due to its demonstrated stability and cost-effectiveness in diverse conversational scenarios~\cite{openai2024gpt4ocard}. 
The responder receives the complete conversation history at each generation step and produces contextually appropriate chatbot responses that are consistent with the established topic and conversational tone. 
This model selection ensures that any variations in dialogue quality observed in our experimental results can be attributed primarily to the inquirer model performance rather than inconsistencies in the responder model's behavior.

\subsubsection{Judge bot}
In the majority of experiments presented in Section~\ref{sec:evaluation}, GPT-4o~\cite{openai2024gpt4ocard} serves as the default judge LLM unless otherwise specified.
However, we acknowledge that the choice of the judge LLM may cause evaluation bias -- for example, the judge LLM might favor dialogues generated using the same LLM as the judge bot itself. We have conducted experiments to verify this using two alternative LLMs acting as judge models: Claude 3.7~\cite{claude} and Gemini 2.0 Flash~\cite{gemini}. The results of these experiments are presented and discussed in Section~\ref{subsubsec:judge_bias}. 

\subsection{Fine-tuning}
\label{subsec:fine-tuning}

To enhance the performance of smaller models, we apply supervised fine-tuning using Low-Rank Adaptation (LoRA)~\cite{hu2022lora} and a carefully curated subset of our synthetic dialogue corpus. This fine-tuning process aims to transfer conversational behaviors and strategies learned from more capable models, thereby improving the ability of smaller models to emulate human-like inquiries while retaining their lightweight deployment advantages.

The fine-tuning datasets are carefully constructed by extracting and formatting samples from the same dialogue corpora as used for seed prompt extraction: OASST1 and Chatbot Arena (see Section~\ref{subsubsec:seed_datasets}). Specifically, we adopt a cross-dataset fine-tuning strategy: when evaluating a model on the OASST1 dataset, we fine-tune it using data sampled from the Chatbot Arena dataset and vice versa. This approach is designed to improve the generalization of smaller models across diverse dialogue styles and domains.

\begin{lstlisting}[language=json]
{
  "chat_template": "tokenizer",
  "distributed_backend": "ddp",
  "mixed_precision": "fp16",
  "optimizer": "adamw_torch",
  "peft": "true",
  "scheduler": "linear",
  "unsloth": "false",
  "batch_size": "2",
  "block_size": "1024",
  "epochs": "3",
  "gradient_accumulation": "4",
  "lr": "0.00001",
  "model_max_length": "16384",
  "target_modules": "all-linear",
  "merge_adapter": "true"
}
\end{lstlisting}
We use Hugging Face Autotrain to easily generate fine-tuning scripts and here we show the parameters we use. Further details on training configurations and evaluation procedures can be found in our publicly available code repository\footref{repo}.

\section{Evaluation}
\label{sec:evaluation}

This section presents the methodology and results of our experimental analysis of various LLMs used in DialogueForge. First, in Section~\ref{subsec:metrics} we describe the evaluation metrics used to assess and compare the performance of various LLMs. Then, in Section~\ref{subsec:results} we present the experimental results obtained using various LLMs, including those fine-tuned in our study.

\subsection{Evaluation metrics}
\label{subsec:metrics}
We employ two evaluation metrics introduced in BotChat~\cite{duan2023botchatevaluatingllmscapabilities}.

\textbf{UniEval}: In this evaluation, the generated conversation is independently assessed by the LLM judge, which determines whether the dialogue resembles a genuine human-chatbot interaction. For each conversation, the LLM judge performs the following steps:
\begin{enumerate}
\item Determines whether the conversation appears to involve an AI participant (Yes/No).
\item If the answer is "Yes," identifies the first utterance it recognizes as AI-generated and outputs its index.
\item Provides a rationale for its decision.
\end{enumerate}
Below is a sample UniEval result.

\begin{lstlisting}[language=json]
{
  "choice": "Yes",
  "index": "3",
  "reason": "The human utterance in the third chat is overly detailed and structured, resembling a list of questions that are too perfectly organized and comprehensive for a natural human conversation. It lacks the informal, sometimes inconsistent phrasing typical of human dialogue, and instead reads like a scripted set of questions designed to cover all aspects of the topic systematically."
}
\end{lstlisting}
\textbf{GTEval}: In this setting, the LLM judge is presented with a pair of dialogues, one of which is an authentic conversation between a human and a chatbot. The judge is tasked with identifying whether either dialogue appears to involve AI-generated content. It also provides a rationale for its decision.

GTEval plays a crucial role in the evaluation framework as it assesses whether the generated conversations are indistinguishable from real human-chatbot interactions.

A sample GTEval result is shown below:

\begin{lstlisting}[language=json]
{
  "choice": "Conversation 2",
  "reason": "In Conversation 2, the human utterances are unusually detailed and structured, with a level of fluency and coherence that is atypical for spontaneous human conversation. The human responses include multiple questions and reflections in a single turn, which is more characteristic of AI-generated text. In contrast, Conversation 1 has more typical human-like brevity and directness in the questions asked."
}
\end{lstlisting}

Both evaluation metrics are designed to assess authenticity and human-likeness of conversations, addressing the primary interest of our study. They do not encompass other dimensions of dialogue quality assessment, such as task completion rates, factual accuracy, or toxicity, which fall outside the scope of our evaluation framework.

\subsection{Experimental results}
\label{subsec:results}

In this section, we present the results of all experiments conducted to evaluate DialogueForge. First, in Sections~\ref{subsubsec:results_unieval} and \ref{subsubsec:results_gteval} we present the UniEval and GTEval scores, respectively, achieved by various inquirer LLMs considered in DialogueForge. In Section~\ref{subsubsec:fine-tuning} we study the performance improvement achieved with smaller LLMs thanks to supervised fine-tuning. Then, in Section~\ref{subsubsec:long_dialogue} we examine how model performance degrades as conversation length increases. Finally, in Section~\ref{subsubsec:judge_bias} we verify the impact that the choice of the judge LLM has on the final evaluation scores.

\subsubsection{UniEval Results}
\label{subsubsec:results_unieval}
Figure~\ref{fig:combined_uni_eval} presents the UniEval passing rates achieved by various inquirer LLMs considered in DialogueForge.
Based on these results, we make the following observations:

\begin{figure}[t]
    \centering

    \begin{subfigure}[t]{\columnwidth}
        \centering
        \includegraphics[width=\textwidth]{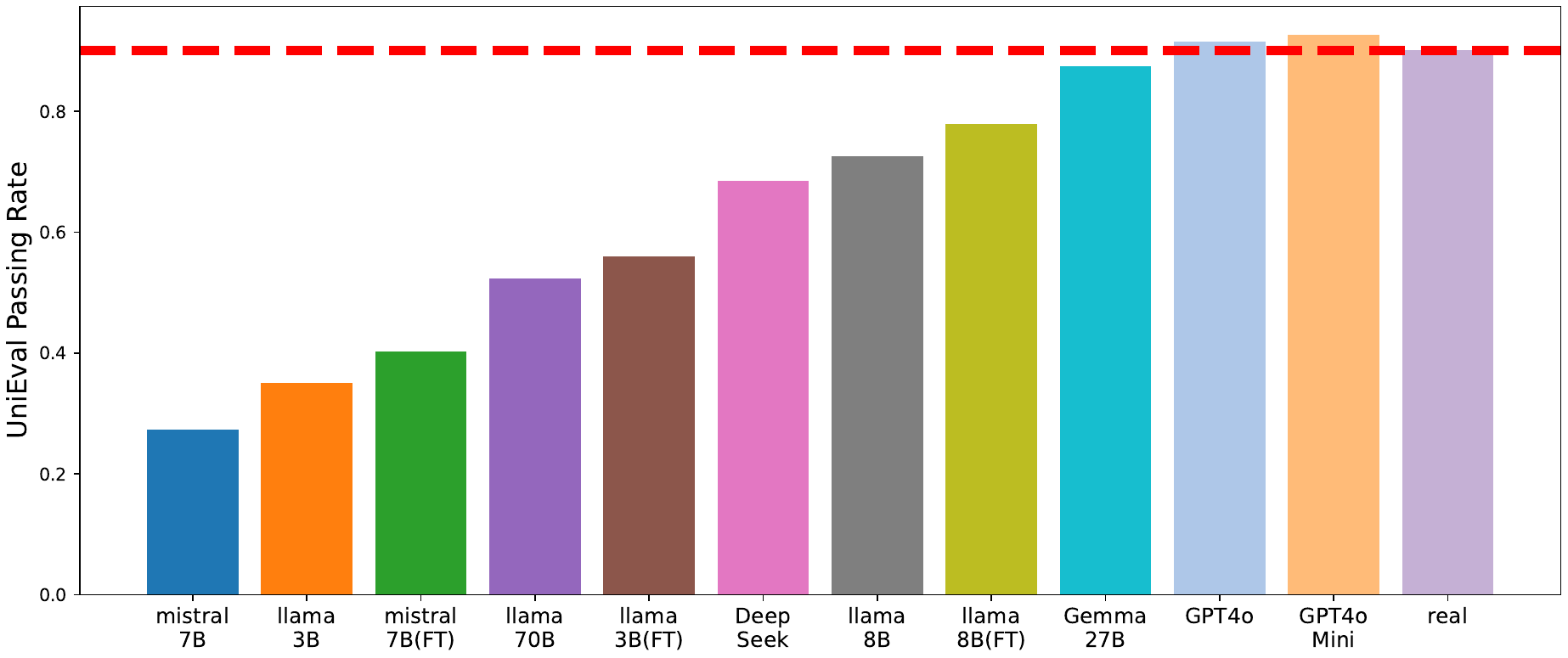}
        \caption{OASST1 dataset: 6 turns}
        \label{fig:oasst_uni_eval_6_turns}
    \end{subfigure}
    \hfill
    \begin{subfigure}[t]{\columnwidth}
        \centering
        \includegraphics[width=\textwidth]{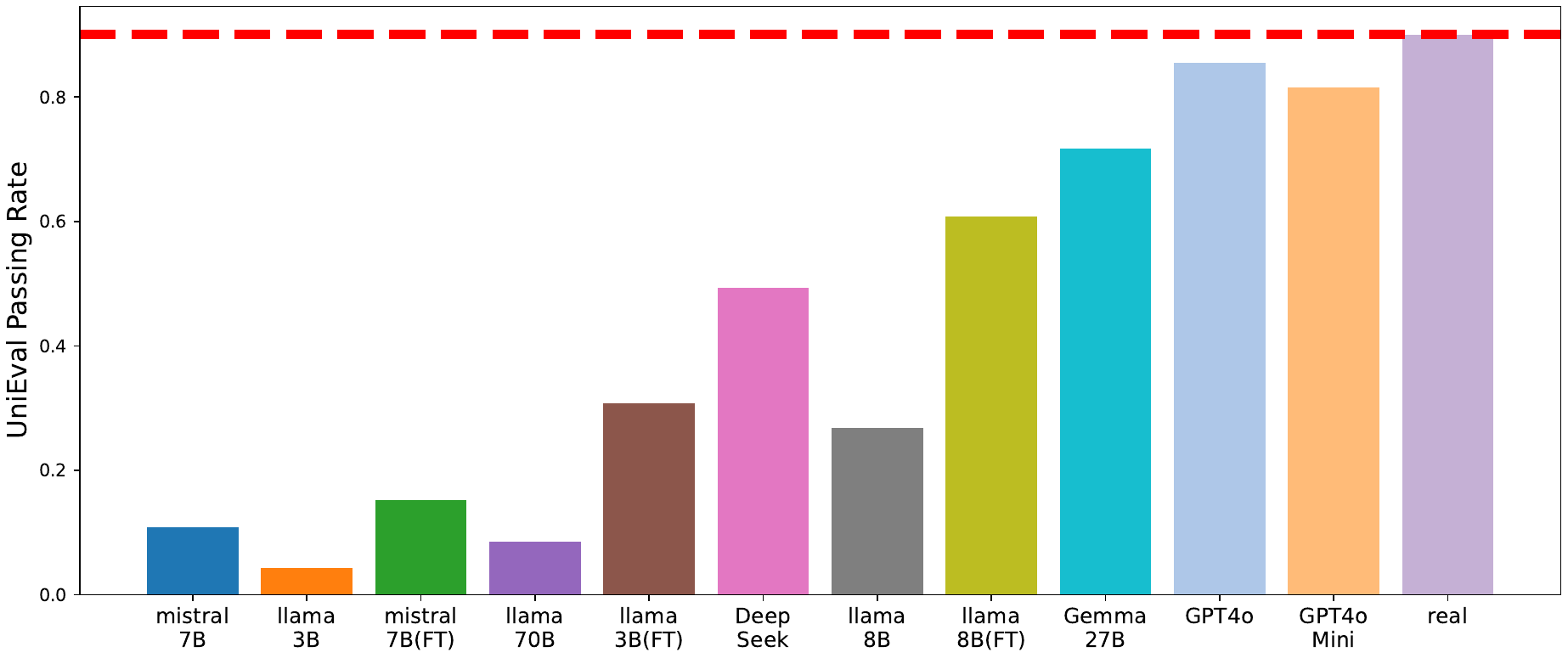}
        \caption{OASST1 dataset: 12 turns}
        \label{fig:oasst_uni_eval_12_turns}
    \end{subfigure}
    \hfill
    \begin{subfigure}[t]{\columnwidth}
        \centering
        \includegraphics[width=\textwidth]{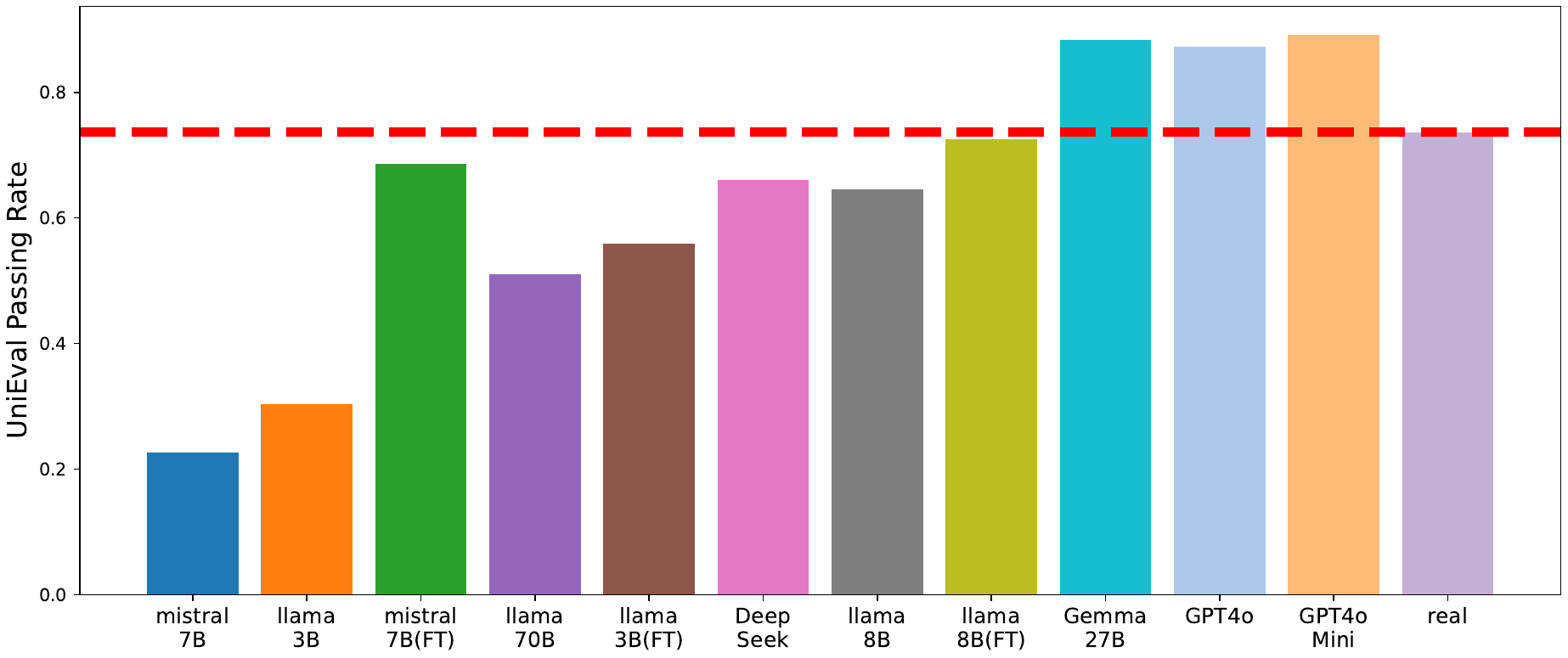}
        \caption{Chatbot Arena dataset: 6 turns}
        \label{fig:arena_uni_eval_6_turns}
    \end{subfigure}
    \hfill
    \begin{subfigure}[t]{\columnwidth}
        \centering
        \includegraphics[width=\textwidth]{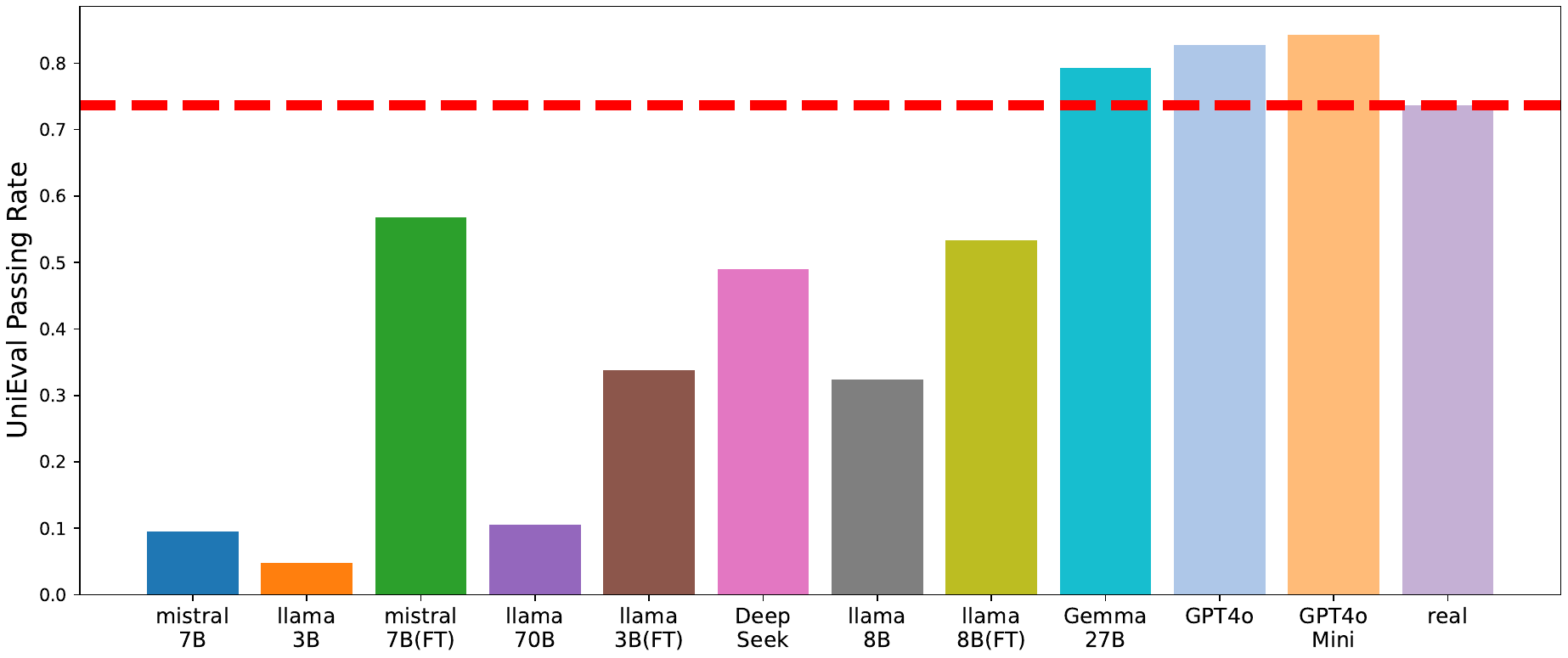}
        \caption{Chatbot Arena dataset: 12 turns}
        \label{fig:arena_uni_eval_12_turns}
    \end{subfigure}

    \caption{UniEval passing rates (No AI-Involved) of OASST1 dialogues. Models labelled with "FT" are fine-tuned. The red dashed line marks the passing rate of real human-chatbot conversations for reference.}
    \label{fig:combined_uni_eval}
\end{figure}

1. \textbf{Larger models generally perform better}. Among the evaluated LLMs, GPT-4o and GPT-4o mini achieve the highest passing rates. On the OASST1 6-turns dialogue dataset, both of them exceed 90\%, even slightly outperforming real human-chatbot conversations. In contrast, smaller models perform significantly worse. For example, on the OASST1 dataset, Mistral-7B achieves a passing rate of only 27.36\%, and Llama-3.2-3B reaches 35.11\%. However, 
\textbf{some exceptions} exist. Notably, Llama-3.3-70B underperforms compared to Llama-3.1-8B, contrary to expectations based on model size. Conversely, Gemma-2-27B performs remarkably well -- its passing rate on the OASST1 6-turn conversation dataset reaches 87.41\%, closely approaching the performance of GPT-4o and GPT-4o mini.

2. \textbf{Fine-tuning improves performance}. All smaller models (e.g. Mistral-7B, Llama-3.2-3B, and Llama-3.1-8B) exhibit improved passing rates after fine-tuning. This demonstrates the effectiveness of task-specific adaptation, even for relatively limited model capacities. A more detailed comparison of fine-tuned versus base model performance is presented in Section~\ref{subsubsec:fine-tuning}.

3. \textbf{LLMs’ ability to simulate human-like dialogue declines with longer conversations}. As the number of utterances in a conversation increases, the passing rates tend to decrease across all models. However, some LLMs exhibit a less pronounced performance drop compared to others, indicating better robustness in maintaining human-likeness over extended interactions. A more detailed analysis of this trend is provided in Section~\ref{subsubsec:long_dialogue}.

\subsubsection{GTEval Results}
\label{subsubsec:results_gteval}
\begin{figure}[t]
    \centering

    \begin{subfigure}[t]{\columnwidth}
        \centering
        \includegraphics[width=\textwidth]{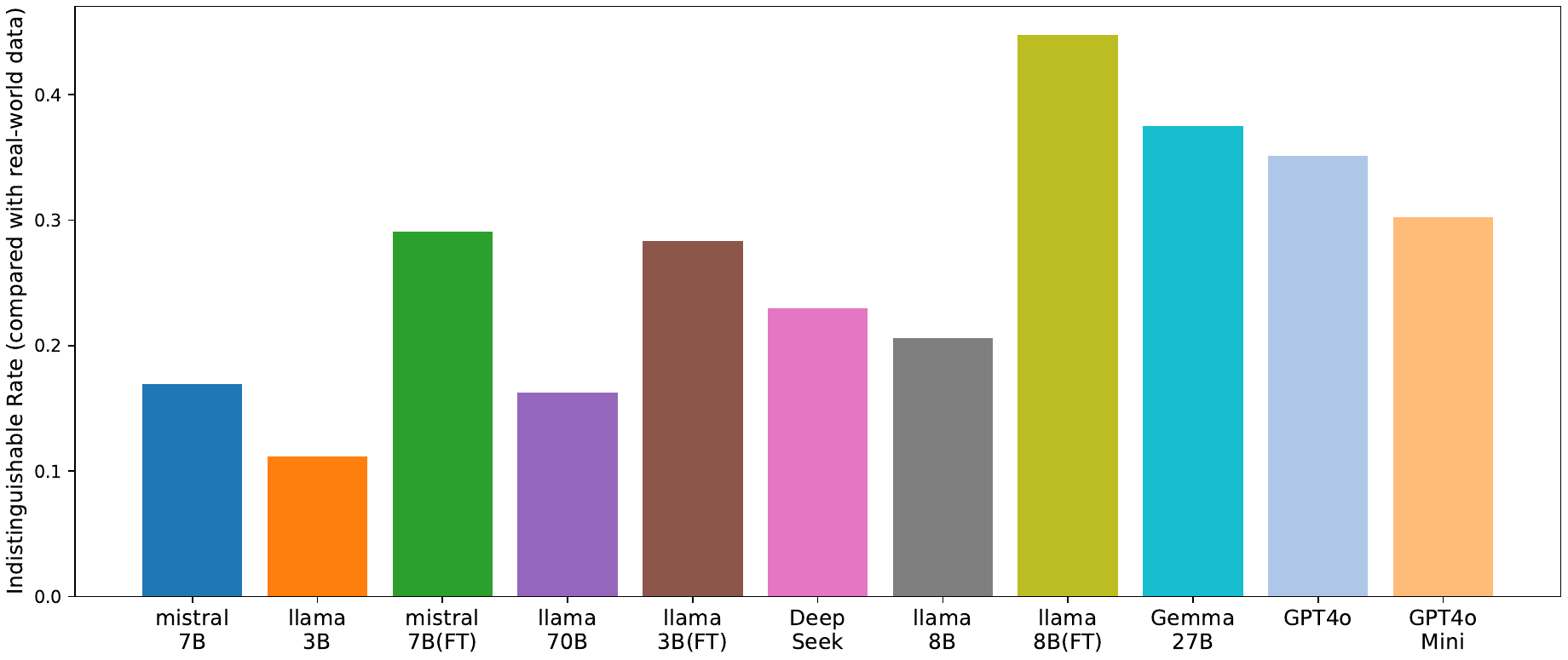}
        \caption{OASST1 dataset: 6-turns}
        \label{fig:oasst_gt_eval_6_turns}
    \end{subfigure}
    \hfill
    \begin{subfigure}[t]{\columnwidth}
        \centering
        \includegraphics[width=\textwidth]{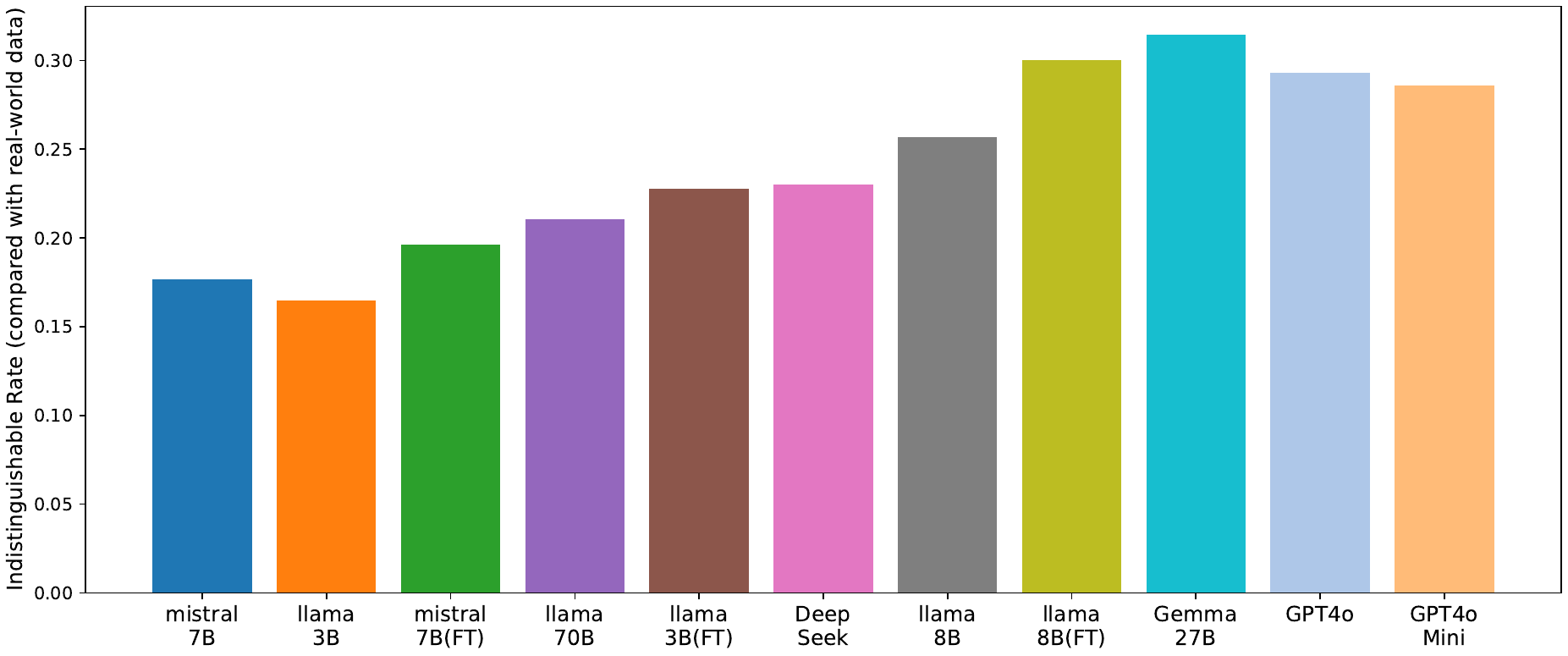}
        \caption{OASST1 dataset: 12-turns}
        \label{fig:oasst_gt_eval_12_turns}
    \end{subfigure}
    \hfill
    \begin{subfigure}[t]{\columnwidth}
        \centering
        \includegraphics[width=\textwidth]{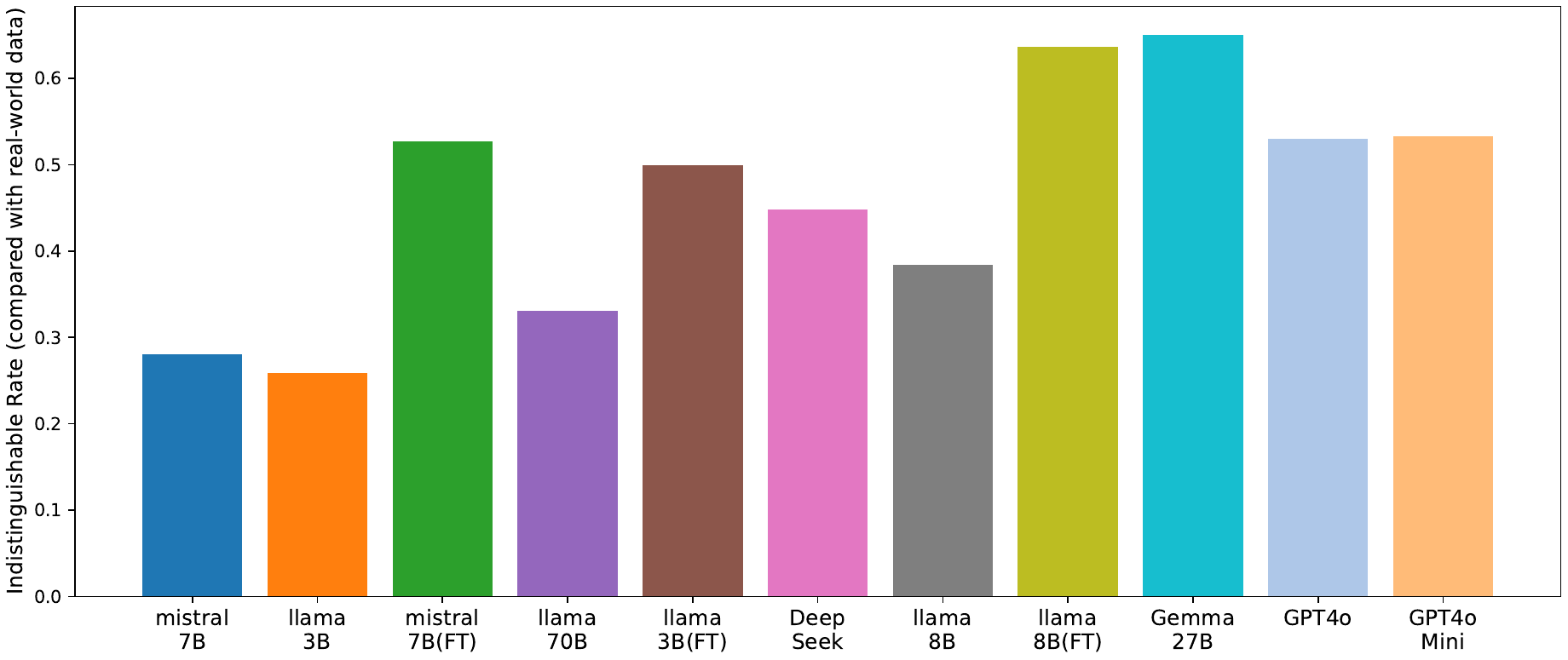}
        \caption{Chatbot Arena dataset: 6-turns}
        \label{fig:arena_gt_eval_6_turns}
    \end{subfigure}
    \hfill
    \begin{subfigure}[t]{\columnwidth}
        \centering
        \includegraphics[width=\textwidth]{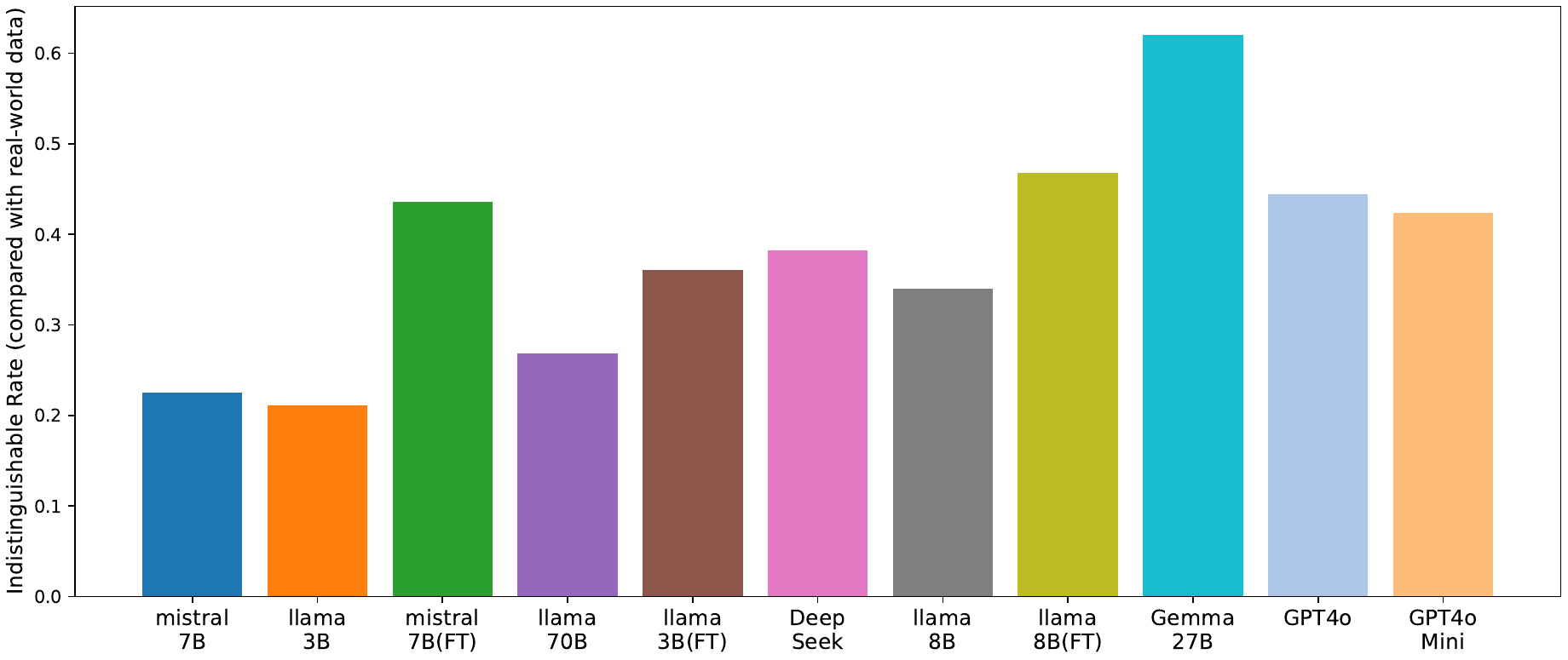}
        \caption{Chatbot Arena dataset: 12-turns}
        \label{fig:arena_gt_eval_12_turns}
    \end{subfigure}

    \caption{GTEval indistinguishability rates (with real data) of OASST1 dialogues. Models labelled with "FT" are fine-tuned.}
    \label{fig:combined_gt_eval}
\end{figure}

Comparing the GTEval results (presented in Figure~\ref{fig:combined_gt_eval}) with the UniEval results (Figure~\ref{fig:combined_uni_eval}), we find that the two evaluation methods generally produce consistent rankings and trends. However, several noteworthy differences and insights emerge, which merit further discussion:

1. \textbf{The GTEval indistinguishability rate is significantly lower than the UniEval passing rate}. This suggests that when a real human-chatbot conversation is provided as a reference point, the LLM judge becomes more discerning in identifying AI-generated dialogues.

2. \textbf{The fine-tuned Llama-3.1-8B model exhibits surprisingly strong performance}. On the OASST1 dataset, for dialogues with 6 utterances, it even achieves a higher GTEval indistinguishability score than GPT-4o. This finding highlights the potential of smaller models to approximate human-like conversational behavior when fine-tuned effectively. It provides evidence that, with targeted fine-tuning, compact LLMs can generate high-quality, realistic dialogues with chatbot that are competitive with much larger models.

\subsubsection{Fine-tuning effects}
\label{subsubsec:fine-tuning}

In this project, we fine-tuned three small-scale language models to assess whether fine-tuning enhances their ability to simulate human-like dialogue in conversations with chatbots. The goal is to evaluate whether such adaptations make smaller models viable for high-quality interaction tasks.

Performance improvements resulting from fine-tuning are presented in Tables~\ref{tab:combined_unieval_improvement} and \ref{tab:combined_gteval_improvement}. We use a cross-dataset evaluation strategy: models fine-tuned on the Chatbot Arena dataset are evaluated on OASST1, and vice versa.

\begin{table}[t]
\centering
\begin{subtable}[t]{\columnwidth}
\centering
\begin{tabular}{llll}
\hline
          & Base (\%) & Finetuned (\%) & Improvement (\%) \\ \hline
Llama-3.2-3B     & 35.11$\mid$4.36  & 55.93$\mid$30.75  & 20.82$\mid$26.39 \\
Llama-3.1-8B     & 72.64$\mid$26.88 & 77.97$\mid$60.77  & 5.33$\mid$33.89  \\
Mistral-7B   & 27.36$\mid$10.90 & 40.19$\mid$15.25  & 12.83$\mid$4.35  \\ \hline
\end{tabular}
\caption{Improvement on the OASST1 dataset (UniEval Passing Rate, No AI-involved). Each cell: 6 and 12 max-turns.}
\label{tab:oasst_unieval_improvement}
\end{subtable}
\hfill
\begin{subtable}[t]{\columnwidth}
\centering
\begin{tabular}{llll}
\hline
          & Base (\%) & Finetuned (\%) & Improvement (\%) \\ \hline
Llama-3.2-3B     & 30.34$\mid$4.79  & 55.88$\mid$33.83  & 25.54$\mid$29.04 \\
Llama-3.1-8B & 64.57$\mid$32.44 & 72.55$\mid$53.29  & 7.98$\mid$20.85  \\
Mistral-7B   & 22.65$\mid$9.48 & 68.56$\mid$56.79  & 45.91$\mid$47.31  \\ \hline
\end{tabular}
\caption{Improvement on the Chatbot Arena dataset (UniEval Passing Rate, No AI-involved). Each cell: 6 and 12 max-turns.}
\label{tab:arena_unieval_improvement}
\end{subtable}
\caption{UniEval Passing Rate improvements after fine-tuning on (a) OASST1 and (b) Chatbot Arena datasets.}
\label{tab:combined_unieval_improvement}
\end{table}

\begin{table}[t]
\centering
\begin{subtable}[t]{\columnwidth}
\centering
\begin{tabular}{llll}
\hline
          & Base (\%) & Finetuned (\%) & Improvement (\%) \\ \hline
Llama-3.2-3B     & 11.14$\mid$16.46  & 28.33$\mid$22.76  & 17.19$\mid$6.30 \\
Llama-3.1-8B     & 20.58$\mid$25.67 & 44.79$\mid$30.02  & 24.21$\mid$4.35  \\
Mistral-7B   & 16.95$\mid$17.68 & 29.06$\mid$19.61  & 12.11$\mid$1.93  \\ \hline
\end{tabular}
\caption{Improvement on the OASST1 dataset (GTEval indistinguishability rates). Each cell: 6 and 12 max-turns.}
\label{tab:oasst_gteval_improvement}
\end{subtable}
\hfill
\begin{subtable}[t]{\columnwidth}
\centering
\begin{tabular}{llll}
\hline
          & Base (\%) & Finetuned (\%) & Improvement (\%) \\ \hline
Llama-3.2-3B     & 25.85$\mid$21.16  & 49.90$\mid$36.03  & 24.05$\mid$14.87 \\
Llama-3.1-8B     & 38.42$\mid$34.03 & 63.67$\mid$46.81  & 25.25$\mid$12.78  \\
Mistral-7B   & 28.04$\mid$22.55 & 52.69$\mid$43.61  & 24.65$\mid$21.06  \\ \hline
\end{tabular}
\caption{Improvement on the Chatbot Arena dataset (GTEval indistinguishability rates). Each cell: 6 and 12 max-turns.}
\label{tab:arena_gteval_improvement}
\end{subtable}
\caption{GTEval indistinguishability rates (with real data) improvements after fine-tuning on (a) OASST1 and (b) Chatbot Arena datasets.}
\label{tab:combined_gteval_improvement}
\end{table}

The benefits of fine-tuning are evident across all three models. Notably, we observe greater performance improvements on the Chatbot Arena dataset (when models are fine-tuned using the OASST1 dataset) than on the OASST1 dataset (when fine-tuned using Chatbot Arena), despite the larger size of the Chatbot Arena dataset. We hypothesize that this discrepancy stems from differences in data quality: the human-chatbot dialogues in the Chatbot Arena dataset may be of lower quality, which could negatively impact the effectiveness of fine-tuning. This highlights the importance of not only dataset size but also the quality and consistency of the training data in model adaptation.

\subsubsection{Performance degradation as the dialogue becomes longer}
\label{subsubsec:long_dialogue}
We analyze how the model performance changes as the length of the conversation increases, focusing on UniEval passing rates and GTEval indistinguishability rates as the number of utterances grows from 6 to 12. The results are summarized in Table~\ref{tab: oasst_rate_change_from_6_to_12}. Based on this data, we make the following observations:

1. \textbf{All models show decreased UniEval passing rates as dialogue length increases from 6 to 12 turns}. This trend suggests that maintaining human-likeness over longer conversations remains a common challenge for all evaluated LLMs. Among the models, GPT-4o, GPT-4o mini, Gemma-2-27B, and DeepSeek-V3 exhibit the highest robustness to increasing dialogue length. In contrast, Llama-3.3-70B shows the most significant decline, with a $43.83\%$ reduction in passing rate -- despite its larger size compared to smaller models like Llama-3.2-3B -- highlighting that model scale alone does not guarantee better performance in long-form dialogue.

2. \textbf{GTEval indistinguishability rates exhibit much smaller changes with increasing dialogue length}. We believe this is because GTEval is already highly effective in detecting AI-generated conversations, even when the dialogues are relatively short.

This performance degradation can be attributed to several factors, including contextual drift, where models lose track of earlier conversation elements, and coherence decay, where responses become increasingly generic or disconnected from the established conversational thread. These phenomena are particularly present in smaller models, which are generally known to exhibit a limited capacity to maintain long-range dependencies across extended dialogue sequences.

\begin{table}[t]
\centering
\begin{tabular}{lll}
\hline
          & UniEval (\%) & GTEval (\%) \\ \hline
Mistral-7B & -16.46    & +0.73    \\
Llama-3.2-3B  & -30.75    & +5.32  \\
Mistral-7B (FT)  & -24.94    & -9.44    \\ 
Llama-3.3-70B  & -43.83    & +4.84    \\ 
Llama-3.2-3B (FT)  & -25.18    & -5.57    \\ 
DeepSeek-V3  & -19.13    & 0.0    \\ 
Llama-3.1-8B  & -45.76    & +5.08    \\ 
Llama-3.1-8B (FT)  & -17.19    & -14.77    \\ 
Gemma-2-27B & -15.74    & -6.05    \\ 
GPT-4o & -6.05    & -5.81    \\ 
GPT-4o mini & -11.14    & -1.69    \\ \hline

\end{tabular}
\caption{The UniEval passing rate and GTEval indistinguishability rate change (+: increase, -: reduction) as the number of utterances in the conversation grows from 6 to 12 (OASST1 dataset).}
\label{tab: oasst_rate_change_from_6_to_12}
\end{table}

\begin{figure}[t]
  \centering
  \includegraphics[width=\linewidth]{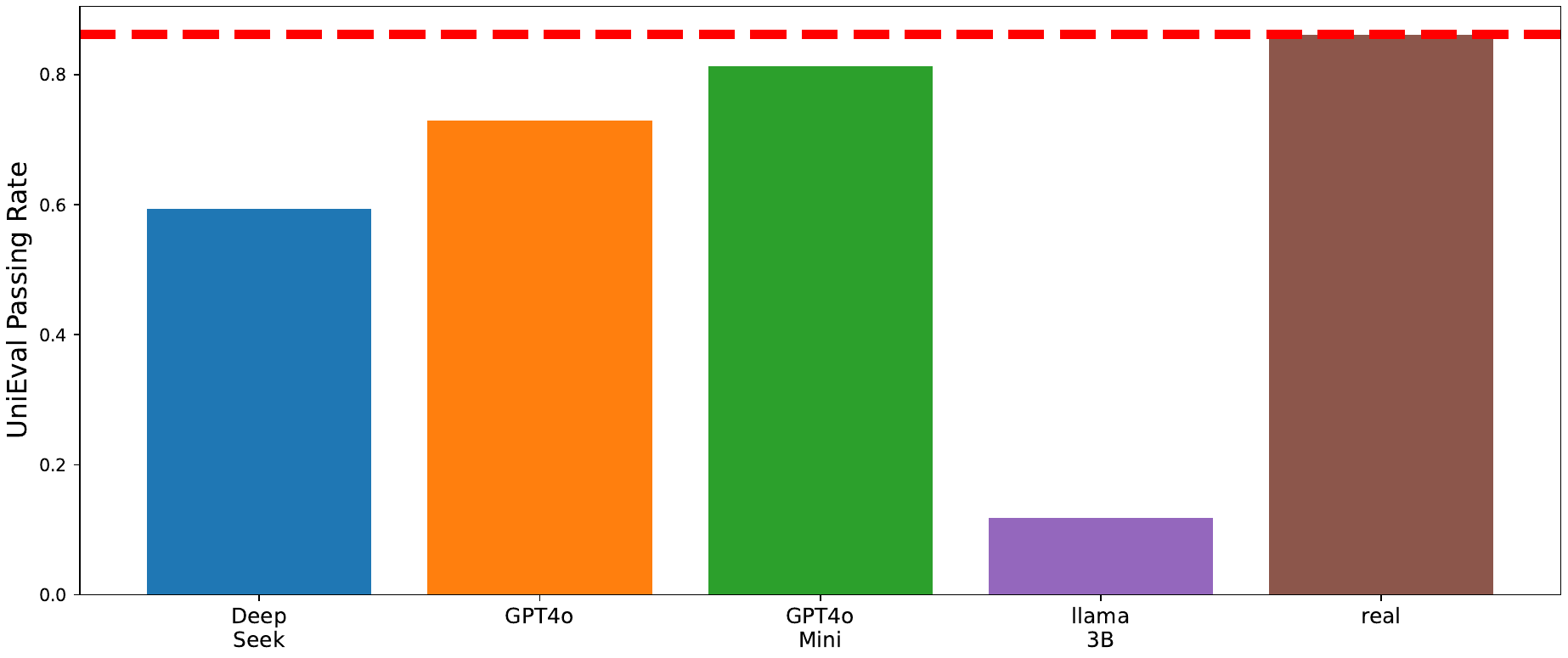}
  \caption{UniEval passing rate (no AI involved) of the OASST1 dataset (6-turns) with Claude 3.7 used as the LLM judge. The red dashed line indicates the passing rate of real human-chatbot conversations.}
  \label{fig:claude_oasst_uni_eval_6_turns}
\end{figure}

\begin{figure}[t]
  \centering
  \includegraphics[width=\linewidth]{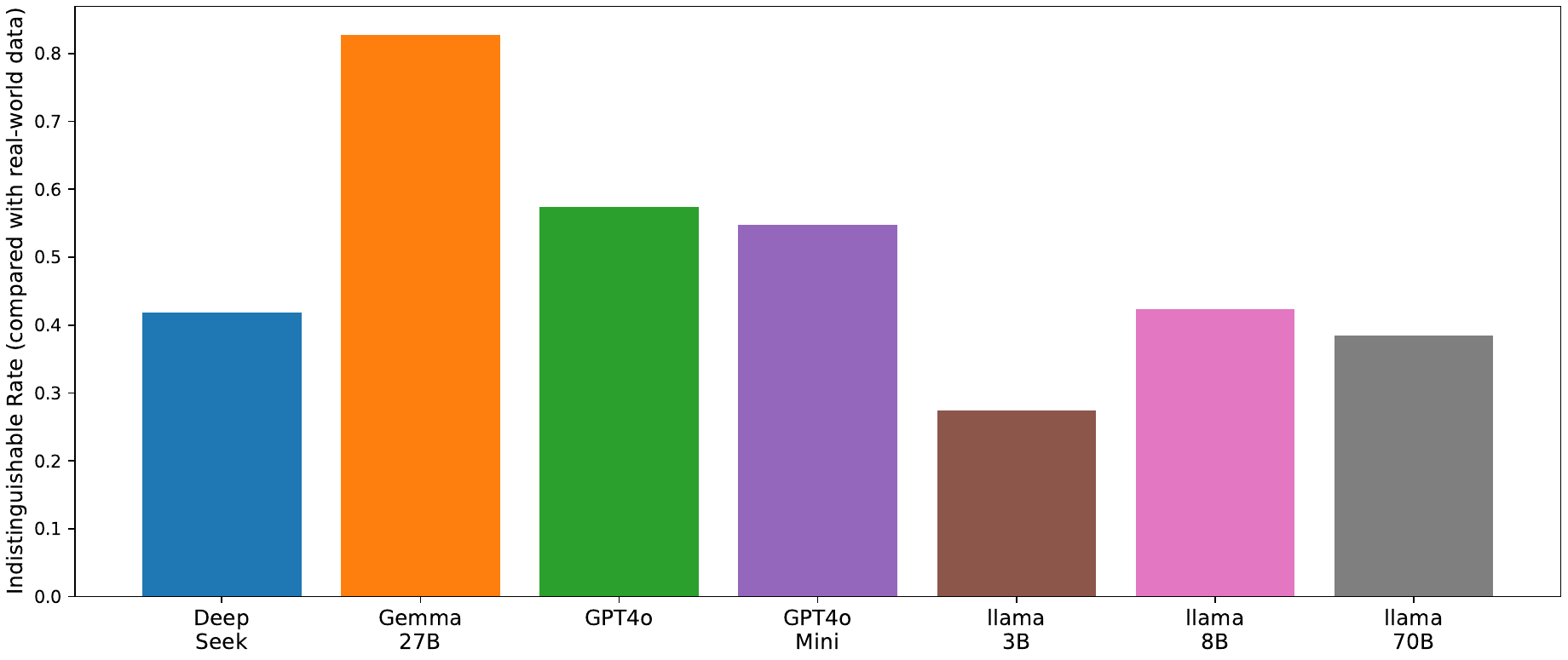}
  \caption{GTEval indistinguishability rate (compared with real data) of the OASST1 dataset (6-turns) with Gemini-2.0-Flash used as the LLM judge.}
  \label{fig:gemini_oasst_gt_eval_6_turns}
\end{figure}

\subsubsection{LLM judge Bias}
\label{subsubsec:judge_bias}
 In this project, GPT-4o is used as the default LLM judge. However, to investigate the possibility of model-specific bias, we conduct additional evaluations using alternative LLM judges: Claude 3.7 and Gemini 2.0 Flash. These models are used to assess dialogues generated by a subset of models. The corresponding results are presented in Figures~\ref{fig:claude_oasst_uni_eval_6_turns} and \ref{fig:gemini_oasst_gt_eval_6_turns}.

Overall, the evaluation results from Claude 3.7 and Gemini 2.0 Flash are largely consistent with those obtained using GPT-4o. When Claude 3.7 is used as the judge LLM (Figure~\ref{fig:claude_oasst_uni_eval_6_turns}), GPT-4o demonstrates a high UniEval passing rate, indicating strong human-likeness. GPT-4o mini also performs well, with a passing rate comparable to that of real human-chatbot dialogues.

When using Gemini 2.0 Flash as the judge LLM (Figure~\ref{fig:gemini_oasst_gt_eval_6_turns}), Gemma-2-27B achieves a surprisingly high GTEval indistinguishability rate. Nonetheless, both GPT-4o and GPT-4o mini continue to show strong performance, outperforming other large models such as DeepSeek-V3 and Llama-3.3-70B.

These results suggest that \textbf{GPT-4o does not exhibit a systematic bias in favor of its own outputs} and can be considered a reliable LLM judge for this evaluation framework.

\section{Discussion}

\subsection{Failure Cases}
In the process of dialogue generation, the inquirer LLM receives only the corresponding prompt as its instruction, resulting in a high degree of output freedom and introducing some unpredictability into the generation process. To prevent infinite turns of conversations, we imposed maximum turn limits for dialogues. The results of the experiments indicate that LLM-generated dialogues tend to have more turns than their real-world counterparts when given specific task summaries, and we have observed that many generated dialogues did not complete the assigned tasks within the given limits. Ideally, the generated dialogues should align closely with the predefined task summaries and follow a specified JSONL format. However, in practice, several unexpected failure cases emerged.

\textbf{Topic deviation.} Certain models, such as Llama-3.2-3B, occasionally generated dialogues entirely unrelated to the assigned task, indicating a significant deviation from the topic of the generation task and reflecting the instability of smaller models in dialogue generation. 

\textbf{Refusal to simulate human-like dialogue.} We provide two prompts to the inquirer, which detail the generation task and instruct it to simulate conversations in human-style. Nevertheless, some models consistently declined to produce simulated human interactions, even under varied prompt settings, thereby halting the generation process. 

\textbf{Discrepancies in output quantity.} Each prompt is derived from a real conversation from the source dataset, ensuring that the number of generated dialogues matches the number of dialogues in the dataset. However, some LLMs unpredictably produced fewer dialogues than expected, and this phenomenon appears to occur in a random and unexpected manner.

\subsection{Limitations}
Our framework enables the scalable simulation of human-chatbot dialogues, but several limitations persist. 
Firstly, while the tested LLMs are capable of generating fluent, grammatically human-like messages, they can fall short in producing goal-oriented dialogues and struggle to maintain long conversation.
This applies to both large pre-trained LLMs and small LLMs improved by fine-tuning. 
As the number of turns increases, the models tend to exhibit contextual drift, loss of coherence by producing generic or irrelevant responses, and lack of task focus.
Secondly, although evaluation with UniEval and GTEval provides useful insights, these metrics may not fully reflect the nuanced qualities of natural conversations, such as the target completion rate or emotional coherence. 
Additionally, our framework does not currently incorporate measures for toxicity or hallucinations, which may limit its applicability in scenarios requiring careful content moderation.
Finally, synthetic dialogues, while fluent, may overfit the stylistic patterns of the models, reducing the diversity typically seen in real human interactions. 
These observations highlight the need for more nuanced evaluation methods and model strategies that go beyond AI-human-involved judgement.

\section{Conclusion \& Future Work}

In this work, we present DialogueForge, a framework for generating human-like multi-turn dialogues using LLMs that simplifies the costly and time-intensive process of collecting human-chatbot dialogue data. 
Unlike existing approaches that focus on single generation paradigms, our framework combines in-context prompting, iterative multi-agent simulation, and supervised fine-tuning to produce diverse, task-tailored multi-turn dialogues at scale. 
As part of our exploration, we employ a variety of both open-source (e.g. Llama, Mistral) and proprietary (e.g. GPT-4o) LLMs of various sizes. 
Our analysis reveals notable differences in human-likeness across models, with proprietary models generally exhibiting stronger coherence and realism, while open-source models offer promising performance with greater controllability and customization. 
Most significantly, we find that the performance gap between large and small models can be substantially reduced through fine-tuning, which improves the performance of smaller language models considerably.

The practical implications of DialogueForge extend beyond the technical contributions. 
By providing an accessible framework for synthetic dialogue generation, our work enables researchers with limited computational resources to contribute to conversational AI research, while publicly available code, models, and datasets lower barriers to entry for dialogue research. 
Our results demonstrate that fine-tuned smaller models can achieve competitive performance, challenging the assumption that only large proprietary models are viable for dialogue generation tasks. 
This finding has important implications for research reproducibility and the development of conversational AI systems in resource-constrained environments.
Overall, we conclude that generating high-quality human-like conversation data using LLMs is possible and can provide satisfactory results depending on the choice of the LLM used to simulate the human inquirer. 
Despite certain limitations of the current results, such as struggles to maintain long conversations, we believe DialogueForge to be a strong contribution towards accessible, scalable dialogue data generation. 

Moving forward, we plan to extend DialogueForge with more complex dialogue scenarios (e.g. persona conditioning) to better reflect real-world interactions and test model robustness under more diverse settings. 
To improve and directly assess models' ability to complete task-oriented dialogues, we also aim to move beyond AI-human-involved judgement by measuring actual goal achievement rates and functional coherence.
Finally, we plan to thoroughly assess our evaluation metrics using human assessments, and employ hybrid evaluation that combines automatic metrics with lightweight human judgement. Such an approach can provide a more comprehensive and reliable assessment of dialogue quality, ultimately accelerating progress toward more natural and effective conversational AI systems.

\bibliographystyle{ACM-Reference-Format} 
\bibliography{bibliography}

\end{document}